\documentclass[a4paper]{article}

\usepackage{INTERSPEECH_v2}
\usepackage{multirow}
\usepackage{makecell}
\usepackage{array}

\newcolumntype{P}[1]{>{\centering\arraybackslash}p{#1}}
\newcolumntype{M}[1]{>{\centering\arraybackslash}m{#1}}

\usepackage{verbatim}

\usepackage{float}
\usepackage{CJKutf8}
\usepackage{adjustbox,rotating}
\title{Order-Preserving Abstractive Summarization for Spoken Content \\ Based on Connectionist Temporal Classification}
\name{Bo-Ru Lu, Frank Shyu, Yun-Nung Chen, Hung-Yi Lee, and Lin-Shan Lee}
\address{College of Electrical Engineering and Computer Science, National Taiwan University, Taiwan}
\email{\{r05942070, b01901051, yvchen, hungyilee\}@ntu.edu.tw, lslee@gate.sinica.edu.tw}

\begin{document}
\maketitle
\begin{abstract}

%With the increasing trend of deep learning, various recurrent neural network based sequence-to-sequence learning models (Seq2Seq) are applied to abstractive summarization by encoding documents into vector representations and then decoding into the summaries.
%However, for Chinese corpus, word-based Seq2Seq models may not perform well on the short summaries (such as headlines), because they often focus on the character level for concise texts.
%In addition, character-level Seq2Seq models do not keep the temporal alignment, so that splitting characters within a word may degrade the summarization performance.
%Connectionist temporal classification (CTC) is a powerful approach for sequence-to-sequence learning, and has been popularly used in speech recognition. The central ideas of CTC include adding a label ``$blank$'' during training. With this mechanism, CTC eliminates the need of segmental alignment, and still maintains the temporal alignments by adding a ``$blank$'' label while decoding. Therefore, this paper proposes to apply CTC to abstractive summarization for spoken content. The ``$blank$'' in this case implies the corresponding input data are less important or noisy; thus it can be ignored. This approach was shown to significantly outperform the state-of-the-art methods in term of ROUGE evaluation over Chinese Gigaword corpus and MATBN corpora. This approach also has the nice property that the ordering of words or characters in the input documents can be better presented in the generated summaries.
Connectionist temporal classification (CTC) is a powerful approach for sequence-to-sequence learning, and has been popularly used in speech recognition. The central ideas of CTC include adding a label ``$blank$'' during training. With this mechanism, CTC eliminates the need of segment alignment, and hence has been applied to various sequence-to-sequence learning problems. In this work, we applied CTC to abstractive summarization for spoken content. The ``$blank$'' in this case implies the corresponding input data are less important or noisy; thus it can be ignored. This approach was shown to outperform the existing methods in term of ROUGE scores over Chinese Gigaword and MATBN corpora. This approach also has the nice property that the ordering of words or characters in the input documents can be better preserved in the generated summaries.
%For Chinese corpus, the methods are usually word-level, because such sequence-to-sequence models do not keep the temporal alignment and splitting characters within a word may degrade the performance.
%However, the short summaries (such as titles) often focus on the character level in order to generate concise texts.
%However, there is no temporal alignment in such models, so that the methods are usually  word-level in order to keep the .

%Connectionist temporal classification (CTC) network is an approach for sequence-to-sequence learning. The main idea of CTC is to add a label ``$blank$'' while decoding. By such mechanism, CTC eliminates the need for forced alignments, and hence is applied to various sequence-to-sequence learning problems. In our work, we applied CTC network to abstractive text summarization. The ``$blank$'' is regarded as unimportant feature when the CTC network reads a noisy word in input data. Our method outperformed existing methods in terms of all ROUGE scores in the Chinese Gigaword corpus.  
\end{abstract}


\section{Introduction}

%Although there are a lot of research on extractive summarization \cite{jing2000sentence}\cite{knight2002summarization}\cite{clarke2008global}\cite{berg2011jointly}\cite{filippova2013overcoming}, the abstractive summarization is an u. 
Spoken content summarization is to generate a summary that describes the core ideas of the spoken document~\cite{tur2011spoken}, which has been performed on several types of spoken content, including news~\cite{maskey2008automatic, lin2009comparative,cai2013ranking}, lectures~\cite{chen2011spoken,shiang2013supervised,lee2014spoken} and meeting records~\cite{xie2009integrating,chen2012two,liu2013towards}.
``Abstractive'' summarization, usually referred to as the ``headline generation'', means that the headline is not necessarily directly extracted from the input spoken document but automatically generated. 
When conducting headline generation \cite{banko2000headline,dorr2003hedge,xu2010keyword,yu2016abstractive}, the model must be capable of both extracting and describing the key concepts of the input. We can formulate this task as a sequence-to-sequence problem, with the machine-transcribed spoken document being the input sequence in the first-stage development, and the summary or headline being the output sentence. This task is important for spoken content, because spoken content is difficult to be shown on the screen, let alone scanned and comprehended by user. Summarization is therefore very helpful for browsing the spoken content. 

Various approaches have been proposed for sequence-to-sequence problems, such as speech recognition \cite{graves2014towards,bahdanau2016end}, neural machine translation~\cite{sutskever2014sequence,bengio2015scheduled}, and dialogue modeling~\cite{shang2015neural, li2016persona}. Among various deep-learning-based methods, Cho \textit{et al.}\cite{cho2014learning} first introduced a basic sequence-to-sequence model based on recurrent neural networks (RNNs) encoder-decoder architecture~\cite{lopyrev2015generating,cheng2016neural,filippova2015sentence,gu2016incorporating}, namely Seq2Seq. An attention mechanism was introduced by Bahdabau \textit{et al.}\cite{bahdanau2014neural}, which allows the decoder to directly fetch the information from input. In Seq2Seq models, an input sequence is encoded into a fixed-length ``sentence vector'' and then decoded into an output sequence. However, this method loses track of the ordering of the elements in the input sequence, since the entire input sequence is encoded into a single ``sentence vector''. In other words, the ordering of the input elements does not directly affect that of the output labels. Order-preserving is desired here, because very often the word order in a spoken document has to do with its semantics. 

This paper proposes a connectionist temporal classification (CTC)~\cite{graves2014towards, graves2006connectionist,graves2013speech,miao2015eesen} based model for spoken content summarization to address the above order-preserving issue. We apply the CTC loss function to impose a hard constraint to preserve the order when generating the output sequence. Unlike the Seq2Seq method that decodes the output sequence after the entire input sequence is read, the CTC loss function decodes the output sequence character by character. That is, the words appearing at the beginning of the input sequence should appear at the beginning of the output sequence if found important. This property of CTC loss function makes it attractive for spoken content summarization. This approach was evaluated on the Chinese Gigaword \cite{graff2005chinese} and the MATBN \cite{wang2005matbn} corpora of Mandarin broadcast news using ROUGE scores metrics \cite{lin2004rouge}.
The results showed that the proposed model significantly outperformed the state-of-the-art approaches \cite{bahdanau2014neural}.

\begin{comment}
\begin{CJK*}{UTF8}{bkai}
\begin{table}
\centering
  \begin{tabular}{ P{2.2cm} p{5cm} }
    \toprule
    \multirow{2}{*}{\makecell{\bf Article} } 
    & 據 印 度 報 業 托 拉 斯 8 日 報 道 印 度 北 方 喜 馬 偕 爾 邦 7 日 夜 發 生 一 起 嚴 重 車 禍 至 少 造 成 36 人 死 亡 16 人 受 傷 當 地 警 方 說 事 故 發 生 在 距 \\
    & A serious car accident occurred in the Himachal Pradesh State, located in the northern part of India, 7 evening, causing at least 36 death and 16 injuries. Indian news Trust reports at 8.\\
    \midrule
    \multirow{2}{*}{\makecell{\bf Ground \\ \bf Truth} }
    & 印 度 發 生 嚴 重 車 禍 至 少 36 人 喪 生\\
    & At least 36 people are killed in a serious car accident in India. \\
    \midrule
    \multirow{2}{*}{\makecell{\bf Proposed \\ \bf Method} } 
    & 印 度 發 生 嚴 重 車 禍 36 人 死 亡\\
    & A serious car accident occurs in India. 36 people died. \\ 
  \bottomrule
  \end{tabular}
  \caption{The performance of the propose method comparing with the ground truth labelled by human. The first row shows the article inputted to the model. The second and third row show the headline labelled by human and generated by the proposed method, respectively. Then original text is in Chinese, the English translation is done by ourselves.}
  \label{table:quick_demo} % from xin200209
\end{table}
\end{CJK*}
\end{comment}

\section{Model}

\begin{figure}[h]
  \includegraphics[width=\linewidth]{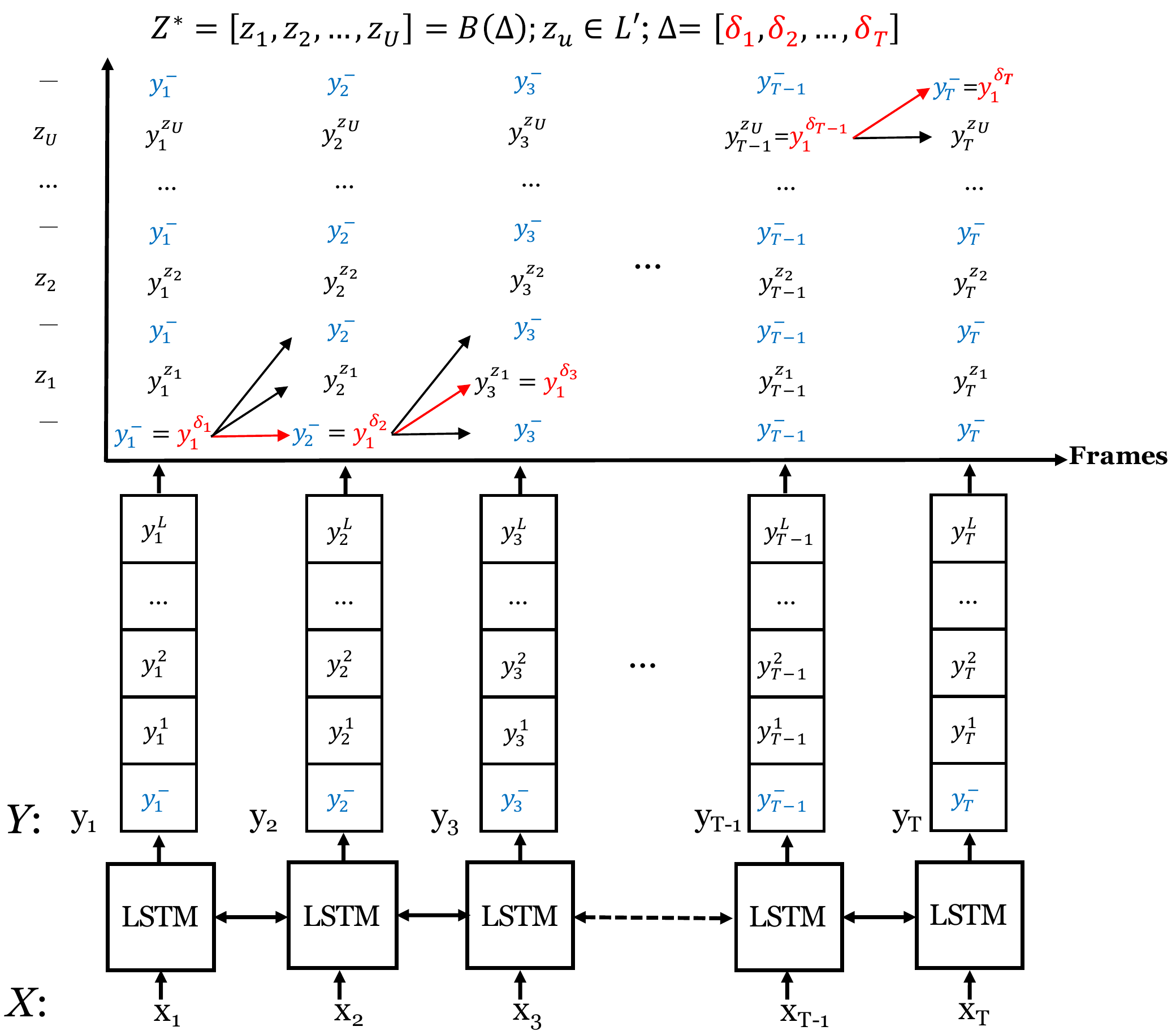}
  \caption{Illustration of the CTC training. For a given input sequence $X=[x_{1},x_{2},...,x_{T}]$ at the bottom, a sequence of vectors $Y=[y_{1},y_{2},...,y_{T}]$ is generated, where $y_{t}$ is a probability distribution over all possible output labels plus the blank ``$-$'', as shown in the lower part of the figure.
The upper part shows a map for finding the probability of a given target sequence $Z^{*}=[z_{1},z_{2},...z_{U}]$ for all possible output sequence hypotheses $\Delta=[\delta_{1},\delta_{2},...,\delta_{T}]$ based on $Y$.
We use the output sequence hypothesis $\Delta$ to interpret here, highlighted in red. All blanks``$-$'' here are highlighted in blue.}
  \label{fig:ctc}
  \vspace{-2mm}
\end{figure}

The proposed model consists of an RNN and a CTC model, where the input is machine-transcribed spoken documents and the output is their summaries.
%In the first-stage development, we formulate the input as characters or words of the machine-transcribed spoken documents and the output as their summaries.
%we select characters or words in machine-transcribed spoken documents and their summaries to be the elements in the input and output sequences. 
The input sequence is segmented into characters or words when feeding into the RNN.
%The input sequence is segmented into such elements before entered into the RNN.
Let $x_t$ denote the input element at time $t$ and the input sequence is then $X=[x_{1},x_{2},...,x_{T}]$, where $T$ is the input sequence length. 
The RNN outputs a probability distribution vector $y_t$ for each time step $t$, so the output sequence is $Y=[y_{1}, y_{2}, ..., y_{T}]$.
%, where $y_{t}$ is a probability distribution vector. 
Finally, the output headline, denoted by $Z=[z_{1}, z_{2}, ...,z_{U}]$, is generated by a forward-backward algorithm that calculates the optimal path based on $Y$. Note that the sequence length of $Z$, denoted by $U$ here, should be significantly less than $T$, because $Z$ is a summary or a headline. Below, we briefly introduce CTC and how it is applied to the summarization task as shown in Figure~\ref{fig:ctc}. 

\subsection{Connectionist Temporal Classification (CTC)}
The connectionist temporal classification (CTC) training is based on a loss function in training RNN, which maps an input sequence $X=[x_{1},x_{2},...,x_{T}]$ onto a target sequence $Z=[z_{1},z_{2},...z_{U}]$.
%Each element $z_{u},u=1,2,...,U$, of the target sequence is an output label, where $z_{u}\in{\{1,2,...,L\}}$ and $L$ is the number of distinct output labels.
The CTC is designed for sequence learning problems that have input and output sequences with different lengths, namely $U \leq T$. For instance, in speech recognition, an acoustic sequence is transcribed into a phoneme sequence. The outputted phoneme sequence is obviously shorter than the acoustic frame sequence.

In CTC training, an extra label ``$blank$'' (denoted by ``$-$'') is inserted between every two adjacent output labels $z_{u}$ and $z_{u+1}$ in the target sequence $Z^*$.
Hence, the output elements have $L'=L+1$ labels, where $L$ is the number of distinct output labels.
Each element in $Y$, $y_{t}$, is an $L'$-dimension vector, $y_{t}=[y^{n}_{t}, n=1,2,...,L']$, where $y^{n}_{t}$ is the probability of observing the $n$-th output label at time index $t$. 
This is shown in the lower part of Figure~\ref{fig:ctc}. Given $Y$, we can then calculate the probability for an arbitrary output label sequence, denoted by $\Delta$.
An output label sequence is regarded as a sequence hypothesis used to generate the final output sequence $Z^*$. This sequence hypothesis is represented by $\Delta=[\delta_{1},\delta_{2},...,\delta_{T}]$, where $\delta_{t}$ is an output label at time index $t$.
$\Delta$ is a sequence whose length equal to that of $Y$, but can be mapped onto the output sequence $Z^*$ with the shorter length by a function $\mathcal{B(\cdot)}$. The function $\mathcal{B(\cdot)}$ is a many-to-one mapping function, which removes the repeated labels and ``$-$'' from ${\Delta}$, but if two identical labels are separated by a ``$-$'', $\mathcal{B(\cdot)}$ generates 2 consecutive labels in the output. For instance, if $\Delta=[\alpha, \alpha, $``$-$''$,$ ``$-$''$, \beta,$ ``$-$''$, \beta]$, then $Z^*=\mathcal{B}(\Delta)=[\alpha, \beta, \beta]$.

The probability of each output sequence hypothesis $\Delta$ can be expressed by
\begin{equation}
\label{eq:probability_of_pi}
P(\Delta \mid X)=\prod_{t=1}^{T} y^{\delta_{t}}_{t},
\end{equation}
where $y^{\delta_{t}}_{t}$ is the probability of observing the output label $\delta_{t}$ at time index $t$ as shown in the upper half of Figure~\ref{fig:ctc}.
During training, given a ground truth target sequence $Z^{*}$, the probability of $Z^{*}$ given $X$ can be calculated by
%where $\forall\delta_{t}\in L'$ and $y^{\delta_{t}}_{t}$ denotes the probability of observing $\delta_{t}$ in time step $t$. From \eqref{eq:probability_of_pi}, the probability of observing a ground truth output sequence $Z^{*}$ given the input sequence $X$ can be calculated by
\begin{equation}
\label{eq:probability_of_z}
P(Z^{*}\mid X)=\sum_{\Delta \in \mathcal{B}^{-1}(Z^{*})} P(\Delta \mid X).
\end{equation}
Because $\mathcal{B(\cdot)}$ is a many-to-one function, many sequence hypotheses $\Delta$ may lead to the same ground truth sequence $Z^{*}$. This is why the summation in \eqref{eq:probability_of_z} is over the set of all such sequence hypotheses denoted by $\mathcal{B}^{-1}(Z^{*})$. The CTC loss function is then defined as in \eqref{eq:loss_function}, %On the upper half of figure~\ref{fig:ctc}, a map constructed for a given $\Delta=[\delta_{1},\delta_{2},...,\delta_{T}]$ is shown, in which a blank ``$-$'' is inserted between every two output elements $\delta_{t-1}$ and $\delta_{t}$. The CTC loss function is then defined as in \eqref{eq:loss_function},
\begin{equation}
\label{eq:loss_function}
\mathcal{L}_{CTC}=-\sum_{\forall(X, Z^{*})\in \theta}\ln P(Z^{*}\mid X),
\end{equation}
where $(X, Z^{*})$ is the pair of the input sequence and its corresponding target sequence, and the $\theta$ denotes the set of all such pairs in the training set. The probability distribution $P(Z^{*}\mid X)$ can be efficiently solved with the forward-backward algorithm. The network can be trained with gradient descent by taking the derivative of \eqref{eq:loss_function}.

\subsection{CTC for Spoken Content Summarization}
In CTC, the blank will be removed in the final decoding stage. Therefore, in the summarization task, if an input element $x_{t}$ has a high probability that $\delta_{t}$ is a blank ``$-$'', the information carried by $x_{t}$ may be less relevant to the core concepts of the document. %Furthermore, the decoding stage ensured that the generated output sequence $Z$ is naturally shorter than the input sequence $X$, eliminating the need of generating an additional mark of ``EOS'' (End-Of-Sentence).

\begin{table*}
\centering
  \caption{Performance comparison of various models and input sequence elements over Chinese Gigaword (no ASR errors).}
  \label{table:comparison}
  \begin{tabular}{|c|c|cccc|cccc|}
    \hline
    \multicolumn{2}{|c|}{\bf Model} & & \bf Input & \bf Output & \bf k & \bf ROUGE-1 & \bf ROUGE-2 & \bf ROUGE-3 & \bf ROUGE-L\\
      \hline\hline
      \multirow{4}{*}{Baseline} &
     \multirow{2}{*}{Seq2Seq} 
     &   (a) & word & character & 1 & 34.47 & 18.30 & 8.82 & 31.26 \\
     & & (b) & character & character & 1 & 36.33 & 18.58 & 8.78 & 32.39 \\
     \cline{2-10}
     & \multirow{2}{*}{Attentive Seq2Seq} 
     &   (c) & word & character & 1 & 36.37 & 20.23 & 10.23 & 32.98 \\
     & & (d) & character & character & 1 & 37.97 & 20.47 & 10.27 & 33.88 \\ 
    \hline
    \multicolumn{2}{|c|}{\multirow{3}{*}{Proposed: CTC} }
     & (e) & word & character & 1 & 25.36 & 9.20 & 3.43 & 24.49 \\
     \multicolumn{2}{|c|}{} & (f) & word & character & 2 & 33.58 & 15.70 & 7.34 & 32.20 \\
     \multicolumn{2}{|c|}{} & (g) & character & character & 1 & \bf 42.71 & \bf 24.62 & \bf 14.24 & \bf 40.56 \\
     \hline
  \end{tabular}
  %\vspace{-3mm}
\end{table*}

\section{Experiments}

%In this section, we report the corpora used and the results. 
%Then, we go through the 5 major experiments conducted. In the first experiment, we optimized our model with respect to parameters $k$ and $N$. Then, the proposed model is compared against existing state-of-the-art models. In the third experiment, we used Longest-Common-Sequence algorithms to prove that time coherence is an important requirement for headline generation tasks. Some results generated by our model are shown in experiment 4 to evaluate the model qualitatively. Last but not least, we tested Attentional Seq2Seq and CTC on spoken content. The result of this section proved that CTC network is capable of generating a good summarization when the there is high LCS score between input article and headline.

\subsection{Experimental Setup}
The experiments are performed on the Chinese Gigaword corpus, which consists of 2.2M of headline-document pairs of news stories covering over 284 months from two Chinese newspapers, namely the Xinhua News Agency of China (XIN) and the Central News Agency of Taiwan (CNA).
We select the headline-document pairs of the first 278 months as the training set, and the remaining 6 months for testing.
This is to test whether the proposed approach is able to predict future headlines given data for the past. This corpus includes only texts but not audio, so the headline-document pairs are good for training. For testing, it is good only for the scenario with zero ASR errors.

In Chinese, a sentence is a word sequence, while a word is composed of one to several characters, and there are no blanks between words in sentences, so a sentence can also be considered as a character sequence.
Therefore, both words and characters can be taken as elements of the input/output sequence.
The Chinese Gigaword corpus is segmented into sequences of words and characters. We compare the performance between using words and characters as input sequence elements below. Before training the model, both word and character vectors were  pretrained with Global Vectors (GloVe)~\cite{pennington2014glove}. The numbers and English words are substituted by tags, which give a word-based vocabulary size of 560,772 and a character-based vocabulary size of 6,752.

To evaluate the summarization of spoken content, we further test the CTC network trained with Chinese Gigaword corpus using the MATBN broadcast news corpus in Mandarin Chinese.
This corpus consists of 198 hours of broadcast news from the Public Television Service Foundation of Taiwan including corresponding manual transcriptions and human-generated headlines. In addition to the manual transcriptions with 100\% accuracy, we also obtain ASR transcriptions using Kaldi toolkit \cite{povey2011kaldi} with a character error rate (CER) of 28.72\%.
The first 126 audio stories are chosen for evaluation. 
For implementing CTC networks, we use 2 layers of bidirectional LSTM \cite{hochreiter1997long,graves2005framewise} and apply the CTC loss function on the top layer of LSTM.
The baselines we compare include the sequence-to-sequence model (Seq2Seq) \cite{cho2014learning} and the attentive Seq2Seq model \cite{bahdanau2014neural}, where we used 1 layer of bidirectional LSTM as the encoder and 1 layer LSTM as the decoder for both Seq2Seq and Attentive Seq2Seq. Other hyperparameters are tuned for optimization of all models.

\subsection{Chinese Sequence Element Preprocessing}

As mentioned above, a Chinese word is composed of one to several characters and a Chinese sentence can be considered as a sequence of words or characters. Very often when a Chinese word is composed of several characters , one of them is the ``head character'' carrying the key information while the others are simply supporting. For example, for the word ``\begin{CJK}{UTF8}{bkai}美國\end{CJK} (the U.S.)'', the first character ``\begin{CJK}{UTF8}{bkai}美\end{CJK}'' is the ``head character'' and this character alone can also carry the same meaning (the U.S.) as a ``mono-character word''. In such cases very often only the ``head character'' is used in the ground truth headline to make it short. On the other hand, many other Chinese words cannot be reduced to ``head characters'' because all characters are important. For example, for the word ``\begin{CJK}{UTF8}{bkai}支持\end{CJK} (support)'', both characters here are important. 

With the above background, it is believed that output sequences for summarization better have characters as their elements, because very often for words with ``head characters'' only the ``head characters'' are used in the ground truth headlines, and we need to be able to either select the ``head characters'' or use the whole word when generating the headlines. For the input sequences, it seems that both words and characters can be elements although characters may be preferred since there are much more distinct words than characters.
Nevertheless, when words are used as the input sequence elements and characters as the output sequence elements, very often the machine only selects a part of those words without ``head characters'' to appear in the output sequences. In order to have those words without ``head characters'', such as ``\begin{CJK}{UTF8}{bkai}支持\end{CJK} (support)'' mentioned above, appear entirely in the output without being cut and partly deleted, it was found that repeating the same word $k$ times when entering them to the CTC network was helpful in generating the whole word in the output. This is referred to as ``k-fold input'' here. More precisely, for an input sequence consisting of N words, $X = [w_{1}, w_{2}, ..., w_{N}]$, the k-fold input for $k=2$ is to enter $X' = [w_{1}, w_{1}, w_{2}, w_{2}, ..., w_{N}, w_{N}]$ to the CTC network.

\subsection{Input Element Comparison} % between Words and Characters as Input Elements}
% over Chinese Gigaword}

We first consider the input sequence elements: words or characters. As mentioned above, when the words are used, the k-fold input could be applied. The results when the Chinese Gigaword corpus (corresponding to no ASR errors) was used for both training and testing are listed in Table ~\ref{table:comparison}. Rows (e)(f) are for words as input sequence elements with $k=1$ and $2$, while row (g) is for characters as the input sequence elements. We can see $k=2$ offered significant improvement (row (f) vs. (e)), while simply using characters is even better (rows (g) vs. (e)(f)).
The finding is the same as the statistical claim that most Chinese words are bi-character words~\cite{huang2015acbima}.
%compared the performance of models using Chinese words ("word-based" models) and Chinese characters ("char-based" models) as input units. This step is to optimize the proposed method before comparing it. One core variable that affects the performance of the word-based models is $k$, since such models have a domain mismatch between the input and output. Recall that $k$ is related to the number of characters in words appearing in headline. To count the number of characters in words, we adopted Jie-Ba toolkit [CITE] to segment the headlines into words. Over 82.6\% of words contained 2 characters or less. Therefore, we chose $k=2$ to perform the K-fold Sequence technique. After applying the K-fold Sequence for word-based model, we achieved a 31.48\% improvement in ROUGE-L score, as shown in row (f)(g) of Table ~\ref{table:comparison}.

%The k-fold input technique was also applied to the case, taking characters as the input sequence elements. However, we found choosing $k=2$ led to a long convergence time while not increasing the ROUGE performance much.

%Overall, the char-based models outperformed the word-based models. The main reason is that the input of word-based models used 300-dim word vectors as input. The information contained in such word vectors is inherently less than the 6,752-dim character vectors used by the char-based model.

\subsection{Comparison with Sequence-to-Sequence Models}

We then compare the performance of the proposed CTC network with two strong baselines, sequence-to-sequence model (Seq2Seq) and the attentive Seq2Seq model. The results are listed on the next two sections of Table~\ref{table:comparison}. We see from Table~\ref{table:comparison}, for both Seq2Seq and Attentive Seq2Seq, the cases where characters as the input sequence elements mostly perform better (row (d) vs. (c), (b) vs. (a)). In addition, the attention mechanism clearly helped (rows (c)(d) vs. (a)(b)). The proposed CTC approach with character as input in row (g) achieves the best summarization performance.%The proposed models were evaluated after they converged. Table~\ref{table:comparison} listed the experiment results.

\begin{table}
\centering
  \caption{Experiment results demonstrating the order-preserving property. The high LCS set accounts for 82.20\% of all documents in the test set. The Attentive Seq2Seq is denoted by Att-Seq2Seq. ROUGE-1, ROUGE-2 and ROUGE-L are denoted by   R-1, R-2 and R-L, respectively.}
  \label{table:PTC}
    \begin{tabular}{|c|cc|ccc|}
    \hline
    \bf Data & &\bf Model & \bf R-1 & \bf R-2 & \bf R-L\\
      \hline\hline
    \multirow{2}{*}{\makecell{High \\82.20\%}} 
    %& \multirow{2}{*}{17.80}
    & (a) & Att-Seq2Seq &     41.12 &     22.66 &     36.73\\
    & (b) &         CTC & \bf 47.09 & \bf 27.83 & \bf 44.91\\
    \hline
    \multirow{2}{*}{\makecell{Low \\17.80\%}}
    %& \multirow{2}{*}{82.20}
    & (c) & Att-Seq2Seq & \bf 23.40 & \bf 10.29 & \bf 20.65\\
    & (d) &         CTC &     22.50 &      9.83 &     20.50\\
    \hline  
  \end{tabular}
\end{table}

\subsection{Investigation of Order-Preserving Property}
CTC networks naturally preserve the ordering of elements from the input sequences and reflect such ordering in the output sequences. Seq2Seq models cannot do so, because they compress the input sequences into fixed length vectors and then decode the vectors. 
In order to analyze how this property benefits the task, we use the longest common subsequence (LCS) \cite{hirschberg1977algorithms} algorithm to evaluate the sequence similarity between the ground truth headline and the input document, giving a real number ranging from 0 to 1. This number is a good indicator of how well the ground truth headline preserves the ordering of the input sequence elements.

We divide the Chinese Gigaword corpus into two groups: one with higher LCS scores and another with lower LCS scores, and then compute their ROUGE scores for two groups.
We calculate the LCS scores of all headline-document pairs and find out that 82.20\% of the pairs have LCS scores higher than $0.4$.
Therefore, $0.40$ is chosen as a threshold for dividing the groups. In Table~\ref{table:PTC}, the group with low LCS scores has an average LCS score of $28.64\%$, while the other has an average LCS score of $71.59\%$. 
The former group that has high LCS has significantly better performance than the latter group (rows(a)(b) vs. (c)(d)).
Within the former group, the proposed CTC network significantly outperforms Attentive Seq2Seq (row (b) vs. (a)), while within the latter group, the proposed CTC network performs slightly worse than Attentive Seq2Seq (row(d) vs.(c)).
Considering that the former group accounts for $82.20$\% of all test set, the CTC network produces results closer to the ground truth headline when the ground truth headline preserves the ordering better, or the CTC network preserves the ordering better.

\begin{CJK*}{UTF8}{bkai}
\begin{table}
\centering
  \caption{An example of generated headlines from various models over an example document in the Chinese Gigaword corpus. The English translation was added here for easy reading.}
  \label{table:example}
\begin{adjustbox}{max width=8cm}
  \begin{tabular}{ m{0.15cm} m{1.2 cm} p{6.8 cm} }
    \toprule
    %\bf Example 1
    %\multirow{2}{*}{\makecell{\bf Article} } 
    %& 據 國 際 傳 真 社 引 述 俄 羅 斯 緊 急 事 務 部 官 員 說 親 俄 羅 斯 的 車 臣 共 和 國 總 部 今 天 遭 到 自 殺 炸 彈 攻 擊 至 少 造 成 四 十 六 人 喪 生 七 十 六 人 受 傷\\
    %& The International Fax Agency quoted the Russian emergency officials and said the pro-Russian headquarters in the Republic of Chechnya today suffered a suicide bomb attack, causing at least forty-six deaths and seventy-six injuries.\\
    %\hline
    %\multirow{2}{*}{\makecell{\bf Ground truth} }
    %& 車臣政府大樓遭炸彈攻擊 UNK 至少四十六人遇害\\
    %& The Chechnya government building suffers bomb attack UNK At least 46 persons were killed\\
    %\hline
    %\multirow{2}{*}{\makecell{\bf Seq2Seq}}
    %& 俄羅斯國家遭炸彈攻擊事件\\
    %& Russian state suffers bomb attack\\
    %\hline
    %\multirow{2}{*}{\makecell{\bf Attentional \\ \bf Sq2Seq}}
    %& 俄羅斯總部遭車臣攻擊\\
    %& Russian headquarters attacked by Chechnya\\
    %\hline
    %\multirow{2}{*}{\makecell{\bf CTC \\ \bf (Proposed Method)}}
    %& 俄車臣總部遭自殺炸彈攻擊四十六人亡\\
    %& Russian Chechnya headquarters suffers suicide bomb attack and caused forty-six death\\  
    
    %\midrule
    %\midrule
    
    %\bf Example 2
    \multirow{1}{*}{(a)} 
    & \multirow{1}{*}{\makecell{\bf Text \\ \bf document} }
    & 美國副國務卿阿米塔吉與日本首相小泉純一郎舉行會談以爭取日本支持美國對伊拉克和北韓的政策 這位美國國務院第二號人物禮貌拜會小泉純一郎會談十五分鐘\ \ (The U.S. Deputy Secretary of State Armitage held meetings with Japanese Prime Minister Junichiro Koizumi to fight for Japan's support about the U.S. policy towards Iraq and North Korea. This second prominent figure of the U.S. State Council visited Junichiro Koizumi for a 15-minute meeting.)\\
    \hline
    \multirow{1}{*}{(b)}
    & \multirow{1}{*}{\makecell{\bf Ground \\ \bf truth}}
    & 美副國務卿訪日本爭取支持對伊拉克政策\ \ (The U.S. Deputy Secretary of State Visited Japan to Fight for Supporting the U.S. Policy toward Iraq)\\
    \hline
    \multirow{1}{*}{(c)}
    & \multirow{1}{*}{\bf Seq2Seq}
    & 美國務卿與美國務卿舉行會談\ \ (The U.S. Secretary of State Held Meetings with the U.S. Secretary of State)\\
    \hline
    \multirow{1}{*}{(d)}
    & \multirow{1}{*}{\makecell{\bf Attentive \\ \bf Seq2Seq}}
    & 美國務卿與北韓舉行會談\ \ (The U.S. Secretary of State Held Meetings with North Korea)\\
    \hline
    \multirow{1}{*}{(e)}
    & \multirow{1}{*}{\makecell{\bf Proposed: \\ \bf CTC}}
    & 美阿米塔日會談支持美對伊拉克北韓政策\ \ (The U.S. Armitage and Japan Held Meetings for Supporting the U.S. Policies toward Iraq and North Korea)\\    
  \bottomrule
  \end{tabular}
  \end{adjustbox}
  \vspace{-3mm}
\end{table}
\end{CJK*}

\subsection{Qualitative Analysis} %記得加LCS Score的比較
From the generated headlines, we observe that sometimes the proposed model generates the headlines including synonyms that do not appear in the input document, although such a phenomenon is not shown in the example of Table~\ref{table:example}.
Table~\ref{table:example} is a selected example of the headlines generated by the best model in row (g) of Table~\ref{table:comparison} for an text document in the Chinese Gigaword corpus.
In the part (e), the proposed CTC network compresses the document into a short sentence very close to the ground truth in the part (b) while very well preserving the ordering of characters in the document. 
The baseline models, Seq2Seq and Attentive Seq2Seq, do not have this capacity as observed in the parts (c)(d).
In the part (c), the word ``\begin{CJK}{UTF8}{bkai}美國務卿\end{CJK} (The U.S. Secretary of State)'' appears repeatedly, while in the part (d) the word ``\begin{CJK}{UTF8}{bkai}北韓\end{CJK} (North Korea)'' appears earlier.
Although in the parts (c)(d) the keywords are well extracted and grammatically correct sentences are generated, these generated headlines in fact give incorrect semantics that differs from the original document due to the limitation of preserving the ordering for both part (c) and (d).
For example, in the part (d) the Attentive Seq2Seq generates a headline ``\begin{CJK}{UTF8}{bkai}美國務卿與北韓會談\end{CJK} (The U.S. Secretary of State Holds Talks with North Korea)'', which means very differently from the original document because of the reversed word order.
In the part (c) the repeated keyword ``\begin{CJK}{UTF8}{bkai}美國務卿\end{CJK} (The U.S. Secretary of State)'' does not make sense. 
On the contrary, the proposed CTC network in the part (e) is capable of preserving the word ordering and offers the correct semantics with a grammatically correct and short sentence. 
Also in the part (e) we see the ``head character'' ``\begin{CJK}{UTF8}{bkai}美\end{CJK} (The U.S.)'' is extracted from the complete word ``\begin{CJK}{UTF8}{bkai}美國\end{CJK} (The U.S.)'' in the original document to be used in the headline, while the complete word ``\begin{CJK}{UTF8}{bkai}支持\end{CJK} (support)'' remains unchanged in the headline because there is no ``head character'' in this word.

\begin{table}
\centering
  \caption{Experiment results for the MATBN corpus. The high LCS data accounts for 96.82 \% of all documents. The Attentive Seq2Seq is denoted by Att-Seq2Seq. ROUGE-1, ROUGE-2 and ROUGE-L are denoted by   R-1, R-2 and R-L, respectively.}
  \label{table:ACOUS}
   \begin{adjustbox}{max width=8cm}
    \begin{tabular}{|c|c|cc|ccc|}
    \hline
    \multicolumn{2}{|c|}{\bf Data} & \multicolumn{2}{|c|}{\bf Model} & \bf R-1 & \bf R-2 & \bf R-L\\
      \hline\hline
    \multirow{4}{*}{\begin{sideways}Manual\end{sideways}}
    & \multirow{2}{*}{\makecell{High \\96.82\%}}
    &  (a) & Att-Seq2Seq &     33.21 &     15.65 &     25.02\\
    && (b) &       CTC & \bf 34.65 & \bf 19.67 & \bf 31.82\\\cline{2-7}
    & \multirow{2}{*}{\makecell{Low \\3.18\%}}
    &  (c) & Att-Seq2Seq &     15.63 & \bf  7.04 &     13.41\\
    && (d) &       CTC & \bf 16.33 &      5.52 & \bf 14.03\\
    \hline
    \multirow{4}{*}{\begin{sideways}ASR\end{sideways}}
    & \multirow{2}{*}{\makecell{High \\96.82\%}}
    &  (e) & Att-Seq2Seq & \bf 30.74 &     14.62 &     23.68\\
    && (f) &       CTC &     30.55 & \bf 16.05 & \bf 27.87\\\cline{2-7}
    & \multirow{2}{*}{\makecell{Low \\3.18\%}}
    &  (g) & Att-Seq2Seq &     14.94 &      3.05 &     11.97\\
    && (h) &       CTC & \bf 17.42 & \bf  7.92 & \bf 17.42\\
    \hline
  \end{tabular}
  \end{adjustbox}
%  \vspace{-6mm}
\end{table}

% ptv 52.76 ptv err 46.38 
% Chinese 63.94
% ptv 83.93 ptv_err 73.90
\subsection{Comparison between Manual and ASR Transcripts}
Here we report the results with spoken documents with the MATBN corpus.
We select the best proposed model and the best baseline from Table~\ref{table:comparison}, where the selected CTC network is from row (g) of the table and the baseline is Attentive Seq2Seq from row(d).
%The CTC network here is the best model trained with Chinese Gigaword corpus in row (c) of Table~\ref{table:comparison} and the best baseline, Attentive Seq2Seq. 
Because this model is actually trained with the first 55 characters of all the documents in Chinese Gigaword, while the news stories in MATBN were much longer, we obtain the headlines for the MATBN news stories in the following way.
We use the first 55 characters of the transcriptions to generate the first headline, then shifted the 55-character window by 5 characters (6-th to 60-th characters) to get the next headline. In this way, we generated a total of 20 headlines using the first 150 characters of each news story. 
We then count the number of characters that appeared in both the generated headline and the first 150 characters in the document for all the 20 headlines. The generated headline maximizing this number is chosen as the final headline.
%This is because, for the MATBN corpus, the LCS score between the first 55 characters and the original document is only 52.76\% and 46.38\% for the manual transcription and the ASR transcription, respectively, while the same condition yields a 63.94\% LCS score for the Chinese Gigaword corpus. 
%To compensate for such phenomenon, we divided each document on the MATBN corpus into 20 sub-documents, and the first 5 characters of each document overlapped the last 5 characters of previous document. For each document, we inputted all 20 sub-documents into both the CTC and the attentive Seq2Seq models, thereby generating 20 headlines each. We counted the number of characters that appeared in both generated headlines and original document. The generated headline with the largest number of same characters was chosen as the final output. 

The results in the upper half of Table~\ref{table:ACOUS} are from manual transcriptions, while the lower half for transcriptions obtained with Kaldi including ASR errors. The test set is also divided into two groups according to their LCS scores evaluated with the manual transcriptions. Here the first group of high LCS scores (above $0.42$) includes $96.82\%$ of the document-headline pairs.
%With results consistent with the previous one,
We find that CTC network significantly outperforms Attentive Seq2Seq for documents with high LCS scores (row (b) vs. (a) and (f) vs. (e)), especially in ROUGE-2 and ROUGE-L, for both manual and ASR transcriptions. The ROUGE scores are also much better in the high LCS group (rows (a)(b) vs. (c)(d) and (e)(f) vs. (g)(h)). Moreover, these results clearly show that CTC works equally well with spoken content with ASR errors, and is also good at capturing the ordering of the input sequence on spoken content data sets. 

%Although the CTC model became slightly worse than Attentive Seq2Seq for the low LCS group in some cases(rows (d) vs. (c)) in ROUGE-2, this group included only $3.18\%$ of the test set.

%In experiment 5, we applied our model to the an additional data set, MATBN. This experiment is aimed to proof that our model can as well be applied to spoken content. The MATBN data set is composed of two subsets: the manual transcription subset and the ASR transcription subset. The manual transcription is human-transcribed; therefore we considered it the performance upper bound for all possible methods. The ASR transcription is recognized by Kaldi toolkit, which resulted in ASR error. If the proposed method can handle the ASR transcription, it is robust enough for real-world user scenarios. Each subset of the data was further divided into two "groups" according to their LCS score, therefore generating 4 "groups" in total.

%Table~\ref{table:ACOUS} shows the performance of the CTC network and that of the Attentional Seq2Seq on all four "groups". Consistent with previous results, we found that CTC network surpassed Attentional Seq2Seq in "groups" with higher LCS scores, especially in ROUGE-L. The results showed that CTC is also good at capturing the time coherence of input sequence on spoken content data sets.

\section{Conclusions}
In this paper, we propose to use a CTC network for abstractive summarization of spoken content. Experiments over Chinese Gigaword and MATBN corpora showed that this approach outperformed the existing sequence-to-sequence learning approaches including that using the attention mechanism. This approach was also shown to be robust with ASR errors, which is a crucial requirement for spoken documents. Overall, the proposed approach is able to preserve the ordering of words/characters in the generated summary, which may be a good reason why it produces better summaries.

% CTC network is proposed for Chinese text summarization problems. The model outperformed both existing Seq2Seq-based models with respect to ROUGE scores. Further experiments showed that the proposed model is especially good at handling the time coherence between the articles and headlines, eliminating the need for an additional attention mechanism. Also, the model is applied to spoken content data and found to have a reasonably robust performance in high LCS score when ASR error is present

%The ISCA Board would like to thank the organizing committees of the past INTERSPEECH conferences for their help and for kindly providing the template files.

\newpage
\bibliographystyle{IEEEtran}

\bibliography{mybib}

\begin{thebibliography}{10}
\providecommand{\url}[1]{#1}
\csname url@samestyle\endcsname
\providecommand{\newblock}{\relax}
\providecommand{\bibinfo}[2]{#2}
\providecommand{\BIBentrySTDinterwordspacing}{\spaceskip=0pt\relax}
\providecommand{\BIBentryALTinterwordstretchfactor}{4}
\providecommand{\BIBentryALTinterwordspacing}{\spaceskip=\fontdimen2\font plus
\BIBentryALTinterwordstretchfactor\fontdimen3\font minus
  \fontdimen4\font\relax}
\providecommand{\BIBforeignlanguage}[2]{{%
\expandafter\ifx\csname l@#1\endcsname\relax
\typeout{** WARNING: IEEEtran.bst: No hyphenation pattern has been}%
\typeout{** loaded for the language `#1'. Using the pattern for}%
\typeout{** the default language instead.}%
\else
\language=\csname l@#1\endcsname
\fi
#2}}
\providecommand{\BIBdecl}{\relax}
\BIBdecl

\bibitem{tur2011spoken}
G.~Tur and R.~De~Mori, \emph{Spoken language understanding: Systems for
  extracting semantic information from speech}.\hskip 1em plus 0.5em minus
  0.4em\relax John Wiley \& Sons, 2011.

\bibitem{maskey2008automatic}
S.~R. Maskey, ``Automatic broadcast news speech summarization,'' Ph.D.
  dissertation, Columbia University, 2008.

\bibitem{lin2009comparative}
S.~H. Lin, B.~Chen, and H.~M. Wang, ``A comparative study of probabilistic
  ranking models for chinese spoken document summarization,'' \emph{ACM
  Transactions on Asian Language Information Processing (TALIP)}, vol.~8,
  no.~1, p.~3, 2009.

\bibitem{cai2013ranking}
X.~Cai and W.~Li, ``Ranking through clustering: An integrated approach to
  multi-document summarization,'' \emph{IEEE Transactions on Audio, Speech, and
  Language Processing}, vol.~21, no.~7, pp. 1424--1433, 2013.

\bibitem{chen2011spoken}
Y.~N. Chen, Y.~Huang, C.~F. Yeh, and L.~S. Lee, ``Spoken lecture summarization
  by random walk over a graph constructed with automatically extracted key
  terms,'' in \emph{Twelfth Annual Conference of the International Speech
  Communication Association}, 2011.

\bibitem{shiang2013supervised}
S.~R. Shiang, H.~Y. Lee, and L.~S. Lee, ``Supervised spoken document
  summarization based on structured support vector machine with utterance
  clusters as hidden variables.'' 2013.

\bibitem{lee2014spoken}
H.~Y. Lee, S.~R. Shiang, C.~F. Yeh, Y.~N. Chen, Y.~Huang, S.~Y. Kong, and L.~S.
  Lee, ``Spoken knowledge organization by semantic structuring and a prototype
  course lecture system for personalized learning,'' \emph{IEEE/ACM
  Transactions on Audio, Speech and Language Processing (TASLP)}, vol.~22,
  no.~5, pp. 883--898, 2014.

\bibitem{xie2009integrating}
S.~Xie, D.~Hakkani-T{\"u}r, B.~Favre, and Y.~Liu, ``Integrating prosodic
  features in extractive meeting summarization,'' in \emph{Automatic Speech
  Recognition \& Understanding, 2009. ASRU 2009. IEEE Workshop on}.\hskip 1em
  plus 0.5em minus 0.4em\relax IEEE, 2009, pp. 387--391.

\bibitem{chen2012two}
Y.~N. Chen and F.~Metze, ``Two-layer mutually reinforced random walk for
  improved multi-party meeting summarization,'' in \emph{Spoken Language
  Technology Workshop (SLT), 2012 IEEE}.\hskip 1em plus 0.5em minus 0.4em\relax
  IEEE, 2012, pp. 461--466.

\bibitem{liu2013towards}
F.~Liu and Y.~Liu, ``Towards abstractive speech summarization: Exploring
  unsupervised and supervised approaches for spoken utterance compression,''
  \emph{IEEE Transactions on Audio, Speech, and Language Processing}, vol.~21,
  no.~7, pp. 1469--1480, 2013.

\bibitem{banko2000headline}
M.~Banko, V.~O. Mittal, and M.~J. Witbrock, ``Headline generation based on
  statistical translation,'' in \emph{Proceedings of the 38th Annual Meeting on
  Association for Computational Linguistics}.\hskip 1em plus 0.5em minus
  0.4em\relax Association for Computational Linguistics, 2000, pp. 318--325.

\bibitem{dorr2003hedge}
B.~Dorr, D.~Zajic, and R.~Schwartz, ``Hedge trimmer: A parse-and-trim approach
  to headline generation,'' in \emph{Proceedings of the HLT-NAACL 03 on Text
  summarization workshop-Volume 5}.\hskip 1em plus 0.5em minus 0.4em\relax
  Association for Computational Linguistics, 2003, pp. 1--8.

\bibitem{xu2010keyword}
S.~Xu, S.~Yang, and F.~C.~M. Lau, ``Keyword extraction and headline generation
  using novel word features.'' in \emph{AAAI}, 2010, pp. 1461--1466.

\bibitem{yu2016abstractive}
L.~C. Yu, H.~Y. Lee, and L.~S. Lee, ``Abstractive headline generation for
  spoken content by attentive recurrent neural networks with asr error
  modeling,'' in \emph{IEEE Workshop on Spoken Language Technology}, 2016.

\bibitem{graves2014towards}
A.~Graves and N.~Jaitly, ``Towards end-to-end speech recognition with recurrent
  neural networks.'' in \emph{ICML}, vol.~14, 2014, pp. 1764--1772.

\bibitem{bahdanau2016end}
D.~Bahdanau, J.~Chorowski, D.~Serdyuk, P.~Brakel, and Y.~Bengio, ``End-to-end
  attention-based large vocabulary speech recognition,'' in \emph{Acoustics,
  Speech and Signal Processing (ICASSP), 2016 IEEE International Conference
  on}.\hskip 1em plus 0.5em minus 0.4em\relax IEEE, 2016, pp. 4945--4949.

\bibitem{sutskever2014sequence}
I.~Sutskever, O.~Vinyals, and Q.~V. Le, ``Sequence to sequence learning with
  neural networks,'' in \emph{Advances in neural information processing
  systems}, 2014, pp. 3104--3112.

\bibitem{bengio2015scheduled}
S.~Bengio, O.~Vinyals, N.~Jaitly, and N.~Shazeer, ``Scheduled sampling for
  sequence prediction with recurrent neural networks,'' in \emph{Advances in
  Neural Information Processing Systems}, 2015, pp. 1171--1179.

\bibitem{shang2015neural}
L.~Shang, Z.~Lu, and H.~Li, ``Neural responding machine for short-text
  conversation,'' \emph{ACL}, 2015.

\bibitem{li2016persona}
J.~Li, M.~Galley, C.~Brockett, G.~P. Spithourakis, J.~Gao, and B.~Dolan, ``A
  persona-based neural conversation model,'' \emph{arXiv preprint
  arXiv:1603.06155}, 2016.

\bibitem{cho2014learning}
K.~Cho, B.~Van~Merri{\"e}nboer, C.~Gulcehre, D.~Bahdanau, F.~Bougares,
  H.~Schwenk, and Y.~Bengio, ``Learning phrase representations using rnn
  encoder-decoder for statistical machine translation,'' \emph{arXiv preprint
  arXiv:1406.1078}, 2014.

\bibitem{lopyrev2015generating}
K.~Lopyrev, ``Generating news headlines with recurrent neural networks,''
  \emph{arXiv preprint arXiv:1512.01712}, 2015.

\bibitem{cheng2016neural}
J.~Cheng and M.~Lapata, ``Neural summarization by extracting sentences and
  words,'' \emph{arXiv preprint arXiv:1603.07252}, 2016.

\bibitem{filippova2015sentence}
K.~Filippova, E.~Alfonseca, C.~A. Colmenares, L.~Kaiser, and O.~Vinyals,
  ``Sentence compression by deletion with lstms.'' in \emph{EMNLP}, 2015, pp.
  360--368.

\bibitem{gu2016incorporating}
J.~Gu, Z.~Lu, H.~Li, and V.~O. Li, ``Incorporating copying mechanism in
  sequence-to-sequence learning,'' \emph{arXiv preprint arXiv:1603.06393},
  2016.

\bibitem{bahdanau2014neural}
D.~Bahdanau, K.~Cho, and Y.~Bengio, ``Neural machine translation by jointly
  learning to align and translate,'' \emph{ICLR}, 2015.

\bibitem{graves2006connectionist}
A.~Graves, S.~Fern{\'a}ndez, F.~Gomez, and J.~Schmidhuber, ``Connectionist
  temporal classification: labelling unsegmented sequence data with recurrent
  neural networks,'' in \emph{Proceedings of the 23rd international conference
  on Machine learning}.\hskip 1em plus 0.5em minus 0.4em\relax ACM, 2006, pp.
  369--376.

\bibitem{graves2013speech}
A.~Graves, A.-r. Mohamed, and G.~Hinton, ``Speech recognition with deep
  recurrent neural networks,'' in \emph{Acoustics, speech and signal processing
  (icassp), 2013 ieee international conference on}.\hskip 1em plus 0.5em minus
  0.4em\relax IEEE, 2013, pp. 6645--6649.

\bibitem{miao2015eesen}
Y.~Miao, M.~Gowayyed, and F.~Metze, ``Eesen: End-to-end speech recognition
  using deep rnn models and wfst-based decoding,'' in \emph{Automatic Speech
  Recognition and Understanding (ASRU), 2015 IEEE Workshop on}.\hskip 1em plus
  0.5em minus 0.4em\relax IEEE, 2015, pp. 167--174.

\bibitem{graff2005chinese}
D.~Graff and K.~Chen, ``Chinese gigaword ldc2003t09,'' \emph{Linguistic Data
  Consortium}, 2003.

\bibitem{wang2005matbn}
H.~M. Wang, B.~Chen, J.~W. Kuo, S.~S. Cheng \emph{et~al.}, ``{MATBN}: A
  mandarin chinese broadcast news corpus,'' \emph{International Journal of
  Computational Linguistics and Chinese Language Processing}, vol.~10, no.~2,
  pp. 219--236, 2005.

\bibitem{lin2004rouge}
C.~Y. Lin, ``Rouge: A package for automatic evaluation of summaries,'' in
  \emph{Text summarization branches out: Proceedings of the ACL-04 workshop},
  vol.~8.\hskip 1em plus 0.5em minus 0.4em\relax Barcelona, Spain, 2004.

\bibitem{pennington2014glove}
J.~Pennington, R.~Socher, and C.~D. Manning, ``Glove: Global vectors for word
  representation.'' in \emph{EMNLP}, vol.~14, 2014, pp. 1532--1543.

\bibitem{povey2011kaldi}
D.~Povey, A.~Ghoshal, G.~Boulianne, L.~Burget, O.~Glembek, N.~Goel,
  M.~Hannemann, P.~Motlicek, Y.~Qian, P.~Schwarz \emph{et~al.}, ``The kaldi
  speech recognition toolkit,'' in \emph{IEEE 2011 workshop on automatic speech
  recognition and understanding}, no. EPFL-CONF-192584.\hskip 1em plus 0.5em
  minus 0.4em\relax IEEE Signal Processing Society, 2011.

\bibitem{hochreiter1997long}
S.~Hochreiter and J.~Schmidhuber, ``Long short-term memory,'' \emph{Neural
  computation}, vol.~9, no.~8, pp. 1735--1780, 1997.

\bibitem{graves2005framewise}
A.~Graves and J.~Schmidhuber, ``Framewise phoneme classification with
  bidirectional lstm and other neural network architectures,'' \emph{Neural
  Networks}, vol.~18, no.~5, pp. 602--610, 2005.

\bibitem{huang2015acbima}
T.-H.~K. Huang, Y.-N. Chen, and L.~Kong, ``Acbima: Advanced chinese
  bi-character word morphological analyzer,'' in \emph{ACL-IJCNLP SIGHAN-8},
  2015.

\bibitem{hirschberg1977algorithms}
D.~S. Hirschberg, ``Algorithms for the longest common subsequence problem,''
  \emph{Journal of the ACM (JACM)}, vol.~24, no.~4, pp. 664--675, 1977.

\end{thebibliography}

\end{document}